\documentclass[pdflatex,sn-mathphys-num]{sn-jnl}

\usepackage{graphicx}%
\usepackage{multirow}%
\usepackage{amsmath,amssymb,amsfonts}%
\usepackage{amsthm}%
\usepackage{mathrsfs}%
\usepackage[title]{appendix}%
\usepackage{xcolor}%
\usepackage{textcomp}%
\usepackage{manyfoot}%
\usepackage{booktabs}%
\usepackage{algorithm}%
\usepackage{algorithmicx}%
\usepackage{algpseudocode}%
\usepackage{listings}%
\usepackage{longtable}
\usepackage{placeins}

\theoremstyle{thmstyleone}%
\theoremstyle{thmstyletwo}%

\theoremstyle{thmstylethree}%

\begin{document}

\title[Article Title]{Automated 2D and 3D Segmentation of AMD and DME Lesions in OCT}


\author[1,2]{\fnm{Lucia} \sur{Sundberg}}\email{lc.sundberg@tum.de}

\author[1]{\fnm{Zhihao} \sur{Zhao}}\email{zhihao.zhao@tum.de}

\author*[1]{\fnm{M. Ali} \sur{Nasseri}}\email{ali.nasseri@tum.de}

\affil*[1]{\orgdiv{MAPS Lab}, \orgname{Technical University of Munich}, \orgaddress{\city{Munich}, \country{Germany}}}

\affil[2]{\orgname{University of Wisconsin-Madison}, \orgaddress{\city{Madison}, \state{Wisconsin}, \country{USA}}}


\abstract{Age-related macular degeneration (AMD) and diabetic macular edema (DME) are leading causes of vision loss, and optical coherence tomography (OCT) is the standard modality for detecting and monitoring the subtle lesions that drive treatment decisions. Most deep-learning segmentation work for OCT is validated only in-domain, leaving generalization to clinical data collected under different acquisition protocols largely untested. This work develops and systematically ablates four lesion-segmentation pipelines --- 2D and 3D variants for AMD and DME --- reaching Dice scores of 0.76--0.82 with strong volumetric and surface calibration (r\_vol, r\_surf $\geq$ 0.97 across all four pipelines) on an in-domain validation set. The ablation process establishes a full-volume, calibration-aware adoption standard that catches mechanisms an ordinary slice-level evaluation would keep, and identifies ensemble composition as the most consistent driver of improvement. To test generalization, the models are evaluated on OLIVES, an external clinical cohort with no lesion-level ground truth, using a proxy-metric framework built around biomarker AUROC, central subfield thickness (CST) correlation, and longitudinal concordance. Predictions track clinical biomarkers outside the training distribution, though less strongly than in-domain --- evidence for, not validation of, automated lesion-burden tracking as a clinical tool.}




\maketitle

\section{Background}\label{sec1}

Age-related macular degeneration (AMD) and diabetic macular edema (DME) are two leading causes of vision loss \cite{saha2019automated, almony2023disease}. AMD is a progressive neurodegenerative disease characterized by drusen deposits in the central retina and pigment epithelial detachment (PED) and affects more than 200 million people worldwide \cite{vyawahare2022amd}. DME involves the breakdown of the blood-retinal barrier, leading to retinal vascular leakage and intraretinal fluid (IRF) accumulation, and affects approximately 5.5 percent of individuals with diabetes mellitus \cite{IM20221244}. Left untreated, both can progress to irreversible vision loss, motivating regular image-based monitoring. 

Optical coherence tomography (OCT) is a high-resolution, non-invasive form of biomedical imaging that has become the leading modality for monitoring and diagnosing AMD and DME because it rapidly produces cross-sectional images of retinal tissue structure \cite{FUJIMOTO20009}. OCT generates 2D cross-sections that can be compiled into volumes, allowing for 3D visualization of retinal pathologies such as the PED and IRF lesions present in AMD and DME. Accurate segmentation of these lesions is essential for quantifying disease severity and monitoring treatment response over time \cite{Zhang2025}. 

However, OCT images are inherently affected by speckle noise and low tissue contrast, which can obscure lesion boundaries and make manual segmentation both time-consuming and subject to inter-observer variability. Furthermore, a single OCT volume can contain hundreds of individual B-scans, making manual delineation of lesion boundaries impractical for routine clinical use. This motivates automated deep learning-based segmentation approaches, with U-Net-based architectures emerging as the dominant paradigm for biomedical image segmentation. This architecture involves an encoder-decoder with skip connections; the encoder progressively downsamples to capture context, the decoder upsamples to full resolution, and the skip connections ensure that fine spatial details are retained \cite{ronneberger2015unet}. 2D U-Net architectures process OCT B-scans independently and benefit from pretrained encoders, but sacrifice inter-slice continuity due to a lack of 3D contextual information \cite{Huang2026}. 3D U-Nets capture this volumetric context between slices, but are more computationally expensive to train and lack pretrained encoders. 

Regardless of the specific architecture used, the majority of published OCT segmentation work is developed and evaluated on a single dataset, often from a single imaging device or clinical site \cite{ABDI2025100396}. Therefore, the question remains how well these models generalize to data collected under different acquisition conditions or patient populations.

\section{Introduction}\label{sec2}

Automated segmentation of retinal lesions in OCT images, specifically PED in AMD and IRF in DME, provides a method for faster and more consistent disease monitoring than manual delineation allows. 2D pipelines remain more computationally efficient and are validated more extensively for this task \cite{QuintanaQuintana2025}. However, disease severity and response to treatment are typically assessed by lesion volume across the retina rather than within a single 2D cross-section \cite{MORAES2024100570, You2021}; therefore, 3D segmentation has particular clinical relevance despite being more computationally demanding and less studied than 2D approaches. These lesions are also often small and subtle in early disease stages when detection is most clinically valuable \cite{Wagner, Vidal2025}, further suggesting the need for volumetric segmentation as these ambiguous lesions can be identified more reliably with additional inter-slice context. This paper builds and compares 2D and 3D U-Net-based segmentation pipelines for AMD and DME lesions, including extensive ablation studies, to directly assess these tradeoffs.  Because clinical utility depends on predicted lesion volume tracking true volume, not merely per-voxel overlap, model selection throughout this work evaluates both segmentation accuracy and volume/surface calibration. The resulting 3D pipeline for the evaluation of DME lesions achieves a pooled Dice score of 0.844 and volume/surface calibration of r=0.998/0.996 on in-domain validation data. 

Whether OCT segmentation models generalize to data collected under different conditions remains largely untested \cite{ABDI2025100396}. This paper addresses this gap by evaluating how well the resulting models generalize to an external clinical cohort, using biomarker presence and clinical label correlation as proxy metrics for segmentation quality in the absence of ground-truth labels.

\section{Methods}\label{sec3}

\subsection{Data}\label{subsec2}
This work uses AMD and DME OCT volumes and their corresponding ground-truth lesion segmentation labels from the Huang et al. dataset \cite{Huang2026}: 62 AMD volumes and 42 DME volumes, each divided into training and validation sets using an 80/20 ratio (seed 42), yielding 49 training and 13 validation volumes for AMD and 33 training and 9 validation volumes for DME.  Figure 1 shows a representative example of each dataset's raw B-scans and their corresponding ground-truth lesion annotations. AMD and DME volumes are processed at native resolution rather than resized, in contrast to the reference dataset paper's own downsampling approach. No held-out test set is used. An earlier attempt to reserve a fixed test set further reduced an already small validation pool --- down to as few as 5 volumes for DME --- and disrupted the iterative model-development workflow; this decision was subsequently reverted. All reported in-domain numbers should therefore be read as validation-set performance rather than held-out generalization estimates. External validation was conducted using the OLIVES dataset \cite{prabhushankar2022olives}, which contains longitudinal B-scan sequences captured across multiple clinical visits per patient. Validation was restricted to the TREX-DME cohort (patients 201--256), as that clinical trial studies the outcomes of patients with DME --- matching the IRF segmentation target of this project --- while the other OLIVES cohort (PRIME) consists of diabetic retinopathy patients without DME. Critically, no ground-truth segmentation labels exist for OLIVES; external validation therefore relies on proxy metrics --- correlating predicted lesion volume with the presence of clinically annotated biomarkers and clinical labels such as central subfield thickness (CST) and best-corrected visual acuity (BCVA) --- rather than direct segmentation accuracy.

\begin{figure}[h!]
\centering
\includegraphics[width=0.7\textwidth]{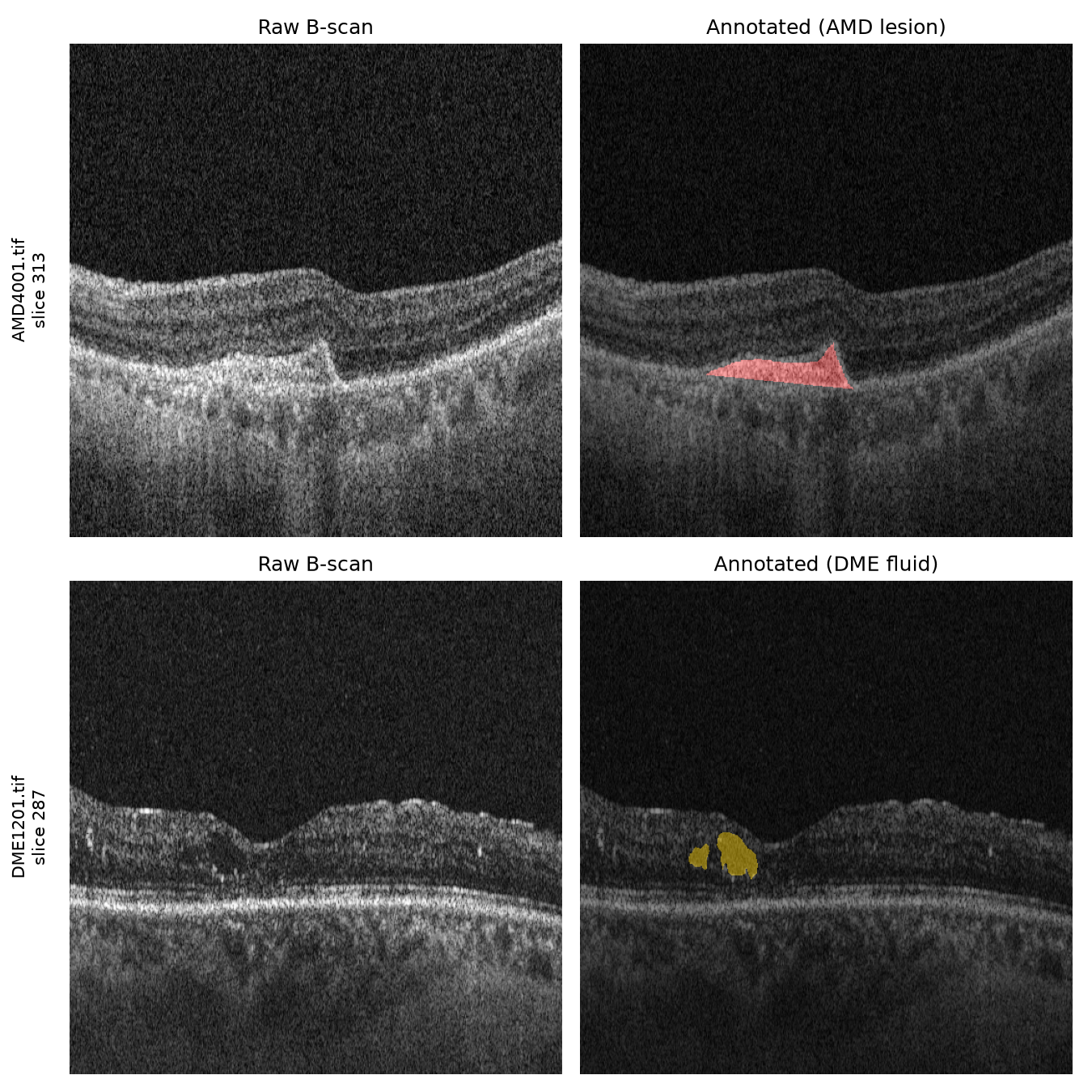}
\caption{Representative OCT B-scans with corresponding ground-truth lesion annotations. Top: AMD lesion (AMD4001, slice 313). Bottom: DME fluid (DME1201, slice 287).}
\label{fig:annotated_bscan}
\end{figure}

\subsection{Model Pipelines}\label{subsec2}

Both the 2D and 3D pipelines use U-Net-based encoder-decoder architectures and are trained using the Adam optimizer with a \verb|ReduceLROnPlateau| learning-rate schedule (factor 0.7, patience 6 epochs), with per-checkpoint variations described below.

The AMD 2D pipeline uses a 3-way ensemble of UNet++ models with a ResNet-34 encoder, combining a baseline model with variants using small-lesion oversampling and weight decay regularization, ensembled uniformly at a decision threshold of 0.55. No post-processing is applied beyond thresholding.

The DME 2D pipeline uses a 4-way ensemble of standard U-Net models with a ResNet-34 encoder, combining a baseline model with variants removing learning-rate decay, using a precision-favoring loss, and using small-lesion oversampling, ensembled uniformly at a decision threshold of 0.45. No post-processing is applied beyond thresholding.

Because 3D models train on fixed-size patches smaller than a full volume, inference uses the MONAI \cite{monai2022} sliding-window approach with Gaussian-weighted blending (overlap 0.25) to reconstruct full-volume predictions.

The AMD 3D pipeline uses a 5-way ensemble of MONAI U-Net models trained from scratch, combining a baseline model with variants using a milder loss at 128-slice patch depth, elongation-aware sampling, full native-depth patches, and small-lesion oversampling, ensembled with weights 0.089/0.089/0.156/0.222/0.444 at a decision threshold of 0.65. No post-processing is applied beyond thresholding.

The DME 3D pipeline uses a 5-way ensemble of MONAI U-Net and AttentionUnet models trained from scratch, combining variants using Dice-BCE loss with weight decay, Dice-BCE loss at 128-slice patch depth, small-lesion oversampling, rotation-only augmentation, and AttentionUnet architecture, ensembled with weights 0.098/0.098/0.098/0.317/0.39 at a decision threshold of 0.60. Predictions are further refined with Gaussian smoothing, a cascade verifier, and a size/texture-based plausibility filter.

The specific ensemble compositions above were selected via a systematic search over training and architectural variants, described in full in the following subsection.

\subsection{Mechanism Catalog}\label{subsec2}

This subsection catalogs every mechanism evaluated across the four segmentation pipelines during model development. A candidate mechanism was adopted into the final composition of a pipeline only if it improved both segmentation accuracy (Dice) and calibration relative to the existing baseline; mechanisms that improved one metric while regressing the other were not adopted. Calibration was assessed as the Pearson correlation between the predicted and ground-truth lesion volume and surface area of each model in the validation set.

\subsubsection{Loss Functions}\label{subsubsec:loss-functions}

Loss function formulation was varied across six configurations per pipeline, spanning standard Dice- and cross-entropy-based losses, class-imbalance-aware reweighting (Tversky, focal Tversky), a boundary-distance penalty, and targeted per-voxel reweighting for small or secondary lesion sites. Table 1 summarizes each variant.
\begin{table}[h!]
\centering
\caption{Loss function variants evaluated across pipelines.}
\label{tab:loss-functions}
\begin{tabular}{p{3.2cm}p{3.0cm}p{6.3cm}}
\toprule
\textbf{Mechanism} & \textbf{Pipeline(s)} & \textbf{Description} \\
\midrule
Dice loss & AMD 2D & Sole soft Dice loss term. \\
Tversky-BCE ($\alpha$/$\beta$ swept 0.2/0.8, 0.35/0.65, 0.6/0.4) & AMD 2D, DME 2D, AMD 3D, DME 3D & Tversky index + BCE; $\alpha$/$\beta$ sets FP/FN penalty balance. \\
Dice-BCE loss & AMD 2D, DME 2D, AMD 3D, DME 3D & Dice + BCE, equal weighting. \\
Boundary-aware distance-transform loss & AMD 2D, DME 2D, AMD 3D, DME 3D & Signed-distance-transform boundary penalty, ramped in over 20 epochs. \\
Focal Tversky loss & AMD 2D, AMD 3D, DME 2D, DME 3D & Tversky term raised to a difficulty-adaptive power. \\
Small-lesion / secondary-component per-voxel reweighting & AMD 2D, DME 2D, AMD 3D, DME 3D& Upweights loss for small/secondary lesion components below a size threshold. \\
\bottomrule
\end{tabular}
\end{table}

\subsubsection{Sampling Strategies}\label{subsubsec:sampling-strategies}

Patch sampling strategy was varied across four mechanisms, addressing positive/negative patch balance, under-representation of highly elongated or small lesions, and synthetic example augmentation via copy-paste. Table 2 summarizes each variant.
\begin{table}[h!]
\centering
\caption{Sampling strategy variants evaluated across pipelines.}
\label{tab:sampling-strategies}
\begin{tabular}{p{4.0cm}p{3.0cm}p{5.5cm}}
\toprule
\textbf{Mechanism} & \textbf{Pipeline(s)} & \textbf{Description} \\
\midrule
Positive-ratio patch sampling (pos\_ratio swept 0.5/0.7/1.0) & AMD 3D, DME 3D (0.7 AMD 3D only) & Fraction of sampled patches guaranteed to contain lesion. \\
Elongation-aware oversampling + full-z-extent positive centering & AMD 3D, DME 3D & 3x patch count for highly elongated lesions; positive centers sampled across full z-extent. \\
Small-lesion volume oversampling & AMD 2D, DME 2D, AMD 3D, DME 3D & 3x patch count for volumes with small lesions. \\
Copy-paste augmentation (lesion-instance and shadow hard-negative variants) & AMD 2D, DME 2D, AMD 3D, DME 3D & Pastes extracted lesion or shadow-artifact crops into training examples. \\
\bottomrule
\end{tabular}
\end{table}

\subsubsection{Patch Depth/Receptive Field}\label{subsubsec:patch-depth}

Receptive field along the z-axis was varied via patch depth and inference-time overlap. Table 3 summarizes each variant.
\begin{table}[h!]
\centering
\caption{Patch depth / receptive field variants evaluated across pipelines.}
\label{tab:patch-depth}
\begin{tabular}{p{4.5cm}p{3.0cm}p{4.8cm}}
\toprule
\textbf{Mechanism} & \textbf{Pipeline(s)} & \textbf{Description} \\
\midrule
Patch-depth sweep (64/128/192/full-native 512 slices) & AMD 3D, DME 3D (192 AMD 3D only) & Z-extent of training patches, from the 64-slice baseline up to the full native volume in one patch. \\
Sliding-window inference overlap increase (0.25 to 0.5) & AMD 3D & Doubles inference-time tile overlap during 3D reconstruction. \\
\bottomrule
\end{tabular}
\end{table}

\subsubsection{Architecture Variants}\label{subsubsec:architecture-variants}

Architecture was varied along three axes: decoder/skip-connection design, encoder backbone, and auxiliary input signal, alongside a deep-supervision variant. Table 4 summarizes each variant.
\begin{table}[h!]
\centering
\caption{Architecture variants evaluated across pipelines.}
\label{tab:architecture-variants}
\begin{tabular}{p{3.8cm}p{3.2cm}p{5.3cm}}
\toprule
\textbf{Mechanism} & \textbf{Pipeline(s)} & \textbf{Description} \\
\midrule
UNet++ (nested/dense skip connections) & AMD 2D, DME 2D, AMD 3D & Nested, densely-connected decoder skip pathways. \\
Decoder-level attention gating & AMD 2D, DME 2D, AMD 3D, DME 3D & Convolutional attention in the decoder — scSE blocks in 2D, MONAI's AttentionUnet gated skip-connections in 3D.\\
EfficientNet-B0 encoder swap & AMD 2D, DME 2D & Replaces ResNet-34 encoder with EfficientNet-B0. \\
ResNet-50 encoder swap & AMD 2D, DME 2D & Replaces ResNet-34 encoder with a deeper ResNet-50. \\
Auxiliary input channels (position / shadow-darkness / combined) & AMD 2D, AMD 3D, DME 2D, DME 3D (shadow and combined: DME 3D only) & Extra input channel(s) encoding anatomical depth position and/or shadow-darkness score. \\
Deep supervision & AMD 3D & Auxiliary loss-supervised outputs at intermediate decoder resolutions. \\
\bottomrule
\end{tabular}
\end{table}

\subsubsection{Augmentation}\label{subsubsec:augmentation}

Augmentation strength was varied from a reduced (rotation-only) recipe up through a standard recipe, and, for domain-generalization testing, intensity perturbations were widened to match a measured gap against an external target domain. Table 5 summarizes each variant.
\begin{table}[h!]
\centering
\caption{Augmentation variants evaluated across pipelines.}
\label{tab:augmentation}
\begin{tabular}{p{4.2cm}p{3.2cm}p{4.9cm}}
\toprule
\textbf{Mechanism} & \textbf{Pipeline(s)} & \textbf{Description} \\
\midrule
Standard augmentation recipe & AMD 2D, DME 2D & Flip, rotation, brightness/contrast, noise, elastic transform, coarse dropout. \\
CoarseDropout removal ablation & AMD 2D & Standard recipe minus coarse dropout. \\
Rotation-only augmentation (reduced recipe) & AMD 2D, DME 2D, AMD 3D, DME 3D & Flip + rotation only. \\
Domain-shift augmentation (full and half strength) & AMD 2D, AMD 3D, DME 2D, DME 3D (half strength: AMD 3D only) & Widened intensity perturbations sized to a measured target-domain gap. \\
Speckle noise injection & DME 3D & Synthetic speckle noise; off by default in production. \\
\bottomrule
\end{tabular}
\end{table}

\subsubsection{Regularization}\label{subsubsec:regularization}

Regularization and optimization stability were varied through weight decay, modifications of the learning-rate schedule, gradient clipping, and extended training budgets. Table 6 summarizes each variant.
\begin{table}[h!]
\centering
\caption{Regularization variants evaluated across pipelines.}
\label{tab:regularization}
\begin{tabular}{p{4.5cm}p{3.0cm}p{4.8cm}}
\toprule
\textbf{Mechanism} & \textbf{Pipeline(s)} & \textbf{Description} \\
\midrule
Weight decay (Adam, \verb|weight_decay=1e-4|) & AMD 2D, DME 2D, AMD 3D, DME 3D & L2 penalty on model weights. \\
LR-decay removal (\verb|--lr-patience 1000|) & AMD 2D, DME 2D & Holds learning rate fixed throughout training. \\
Reduced initial learning rate (1e-4 to 3e-5) & AMD 3D, DME 3D & Stabilization fix for deeper-patch/loss configurations prone to background collapse. \\
Gradient-norm clipping + non-finite-batch guard & AMD 3D, DME 3D & Clips global gradient norm; skips batches with non-finite loss. \\
Extended training epoch budget (100 to 120 epochs) & AMD 3D, DME 3D & Resumes a converged checkpoint for additional epochs. \\
\bottomrule
\end{tabular}
\end{table}

\subsubsection{Post-Processing}\label{subsubsec:post-processing}

Post-processing was concentrated primarily in the 3D pipelines, spanning test-time augmentation and threshold tuning, morphological and connected-component operations, shape- and image-evidence-based filters targeting DME 3D's vessel-shadow artifacts, and two learned component-plausibility models. Table 7 summarizes each variant.
\begin{table}[h!]
\centering
\caption{Post-processing mechanisms.}
\label{tab:postprocessing}
\begin{tabular}{p{4.2cm}p{2.8cm}p{6.5cm}}
\hline
Mechanism & Pipeline(s) & Description \\
\hline
TTA --- flip-based test-time augmentation & AMD 2D, DME 2D, AMD 3D, DME 3D & Averages predictions across flipped inference passes (horizontal; +vertical for 2D; +z-axis for 3D). \\
Decision-threshold tuning & AMD 2D, DME 2D, AMD 3D, DME 3D & Sweeps the binarization threshold on TTA-averaged probabilities. \\
3D binary erosion & AMD 3D, DME 3D & Morphological erosion to remove thin/noisy boundary voxels. \\
3D binary closing + hole-filling & AMD 2D, AMD 3D, DME 3D & Reconnects fragmented components and fills small gaps. \\
Minimum-component-size filtering (absolute and relative) & AMD 3D & Removes components below an absolute or relative size cutoff. \\
Largest-connected-component filtering & AMD 3D & Keeps only the single largest predicted component. \\
Edge-aware guided-filter boundary refinement & AMD 3D & Guided filter using raw image intensity as an edge-aware guide. \\
Component fill-ratio (thin-line artifact) filter & DME 3D & Removes thin/branching components by voxel-to-bounding-box ratio. \\
Component axial-elongation (vessel-shadow-line) filter & DME 3D & Removes tall/narrow components resembling vessel-shadow lines. \\
Image-evidence shadow-suppression filter & DME 3D & Removes components overlapping flagged raw-image shadow columns. \\
Plausibility filter (rule-based thresholds) & DME 3D & Suppresses components via intensity/texture/band-position thresholds. \\
Two-stage cascade / component-verifier network & DME 3D & Small 3D CNN classifies predicted components as keep/suppress. \\
\hline
\end{tabular}
\end{table}

\subsubsection{Ensemble Composition}\label{subsubsec:ensemble-composition}

Ensemble composition and combination weighting were determined through two search procedures. Table 8 summarizes each.
\begin{table}[h!]
\centering
\caption{Ensemble composition variants evaluated across pipelines.}
\label{tab:ensemble-composition}
\begin{tabular}{p{4.5cm}p{3.0cm}p{4.8cm}}
\toprule
\textbf{Mechanism} & \textbf{Pipeline(s)} & \textbf{Description} \\
\midrule
Checkpoint ensembling (uniform and weighted averaging) & AMD 2D, DME 2D, AMD 3D, DME 3D & Combines TTA-corrected member probabilities via uniform averaging or coordinate-ascent-tuned weights. \\
Leave-one-out ensemble composition search & AMD 2D, DME 2D, AMD 3D, DME 3D & Tests dropping/adding members to find the best-performing subset. \\
\bottomrule
\end{tabular}
\end{table}
\FloatBarrier

\subsection{OLIVES Protocol}\label{subsec2}

Because OLIVES B-scans differ substantially in framing from the training-domain data, each B-scan is cropped to the tissue band via a percentile-based row-intensity threshold and resized to 512x512 via bilinear interpolation before inference. Crop tightness is tuned separately per dimensionality (a looser threshold for 2D, a tighter one for 3D), since the tighter setting corrects the tendency for 3D to under-predict but degrades the already-strong 2D signal if applied there. B-scans are grouped by patient, eye, and visit, and restricted to visits with at least 45 scans and no gap greater than one scan index in the sequence, ensuring that each assembled volume represents a nearly complete B-scan sequence. OLIVES does not provide an explicit visit-date or visit-index field; visit boundaries are identified through unique combinations of clinically recorded values (CST, BCVA) for a given patient and eye, and the resulting visit order is inferred from row position in the released data rather than an independently confirmed timestamp. Each visit volume is resampled to match the ensemble's training patch depth, passed through the full adopted DME 3D ensemble with TTA and sliding-window inference, and resampled back to its native depth. The complete adopted post-processing chain (smoothing, thresholding, cascade verifier, plausibility filter) is applied identically to the in-domain pipeline. Because OLIVES exhibits greater cross-domain noise than the training data, predictions are additionally masked against a traced ILM/RPE boundary line --- a guardrail step not used during in-domain evaluation. The DME 2D and AMD 2D pipelines undergo the same crop-and-resize pre-processing and guardrail masking, but operate on individual B-scans directly rather than resampled volumes: each slice is passed through the full adopted 2D ensemble with horizontal-flip TTA, without the depth resampling or sliding-window inference used for the 3D pipeline. DME 2D uses the same intraretinal envelope guardrail as DME 3D; the AMD 2D PED analysis uses the RPE-proximity guardrail variant instead. No fine-tuning or retraining occurs on OLIVES data at any point. Two proxy signals are computed per visit. The predicted lesion volume (total predicted positive voxel count) is correlated (Pearson) with the clinically recorded CST and BCVA of the visit. Separately, per-slice predicted lesion area is paired with per-slice binary presence labels for a set of clinically annotated biomarkers (e.g., intraretinal fluid) to compute AUROC, requiring at least 10 positive and 10 negative slices before a metric is considered reportable. For eyes with two or more visits, directional concordance is also calculated, evaluating whether the change in predicted lesion volume between consecutive visits (increase or decrease) matches the direction of the corresponding change in CST.

\section{Experiments}\label{sec4}

\subsection{Final Model Performance}\label{subsec2}

Table 9 reports the final segmentation and calibration performance across all four pipelines, each evaluated on its own validation set. Figures 2 and 3 show representative case examples for the two 3D pipelines, spanning the range from worst- to best-performing validation volumes.
\begin{table}[h!]
\centering
\caption{Final in-domain segmentation performance for all four pipelines. 2D Dice is computed over positive slices only; 3D Dice is the mean over all validation volumes. Pooled Dice, $r_{vol}$, and $r_{surf}$ are full-volume metrics in all cases.}
\label{tab:final_performance}
\renewcommand{\arraystretch}{1.3}
\begin{tabular*}{\textwidth}{@{\extracolsep{\fill}}lcccc@{}}
\hline
Pipeline & Dice & Pooled Dice & $r_{vol}$ & $r_{surf}$ \\
\hline
AMD 2D & 0.7641 & 0.8451 & 0.9880 & 0.9807 \\
DME 2D & 0.7722 & 0.8490 & 0.9981 & 0.9985 \\
AMD 3D & 0.7623 & 0.7573 & 0.9723 & 0.9927 \\
DME 3D & 0.8247 & 0.8444 & 0.9983 & 0.9963 \\
\hline
\end{tabular*}
\end{table}
\FloatBarrier

\begin{figure}[h!]
\centering
\includegraphics[width=\textwidth]{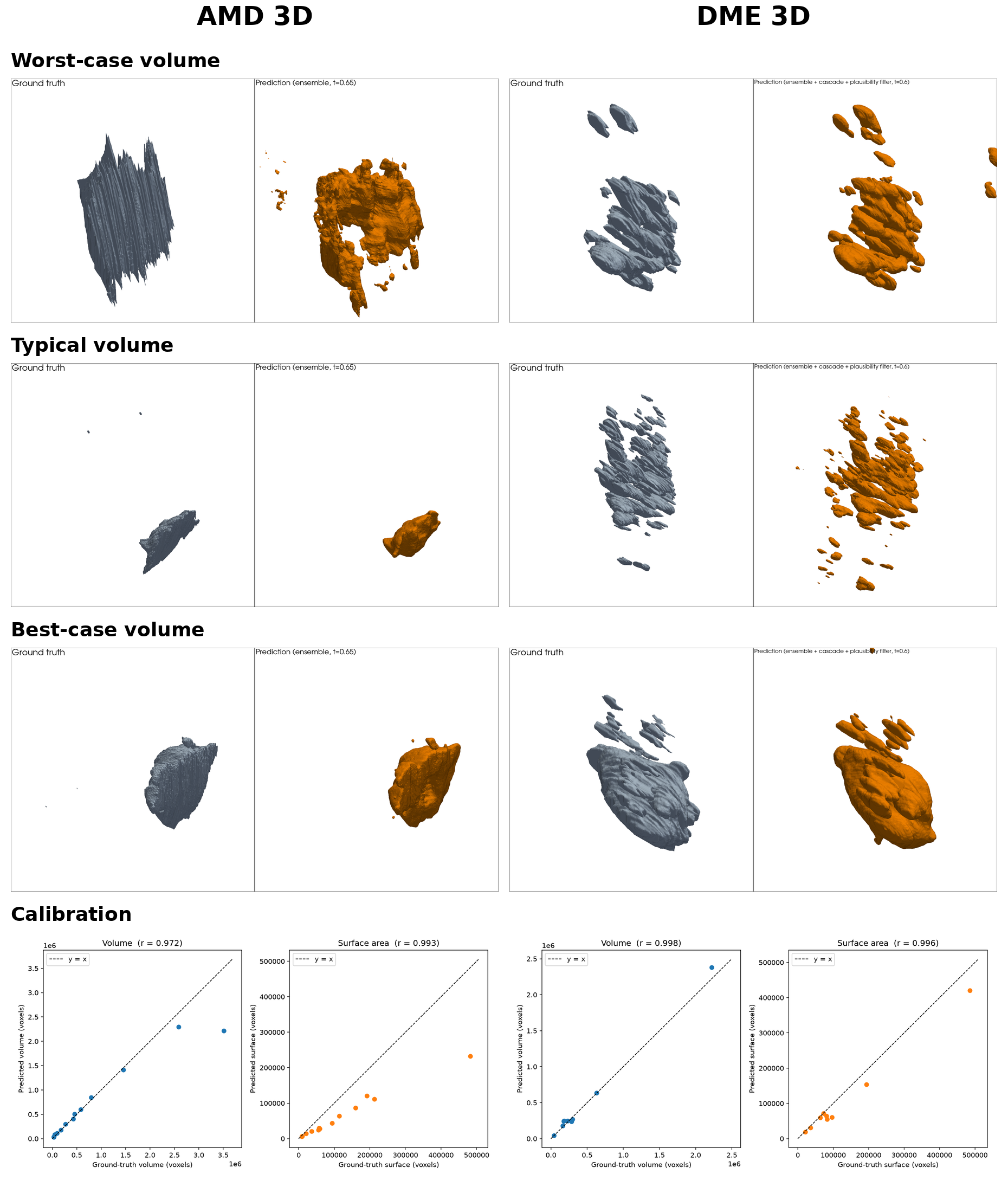}
\caption{AMD 3D and DME 3D case examples (worst-, typical-, and best-case validation volumes by full-volume Dice) and calibration (predicted vs. ground-truth volume and surface area) for each pipeline's current production configuration.}
\label{fig:combined_3d}
\end{figure}

For external context, the Huang et al. [8]baseline 3D U-Net reaches Dice scores of 79.73\% (AMD) and 81.00\% (DME) using this dataset, and their best proposed architecture (a BiFormer-attention 3D network) reaches 82.72\% (AMD) and 86.84\% (DME). These numbers are not directly comparable; Huang et al. evaluate on lesion-centered 384x384x64 sub-volumes cropped and resized from the native 512x512x1044 volume, whereas this project evaluates on complete, native-resolution volumes including lesion-free regions, a more difficult and clinically realistic setting since a deployed model has no oracle indicating where to crop. With that caveat, this project's DME 3D pipeline (82.47\% mean Dice) exceeds the Huang et al. plain 3D U-Net baseline (81.00\%) despite the harder evaluation protocol; AMD 3D (76.23\%) trails their baseline (79.73\%).

\begin{figure}[h!]
\centering
\includegraphics[width=\textwidth]{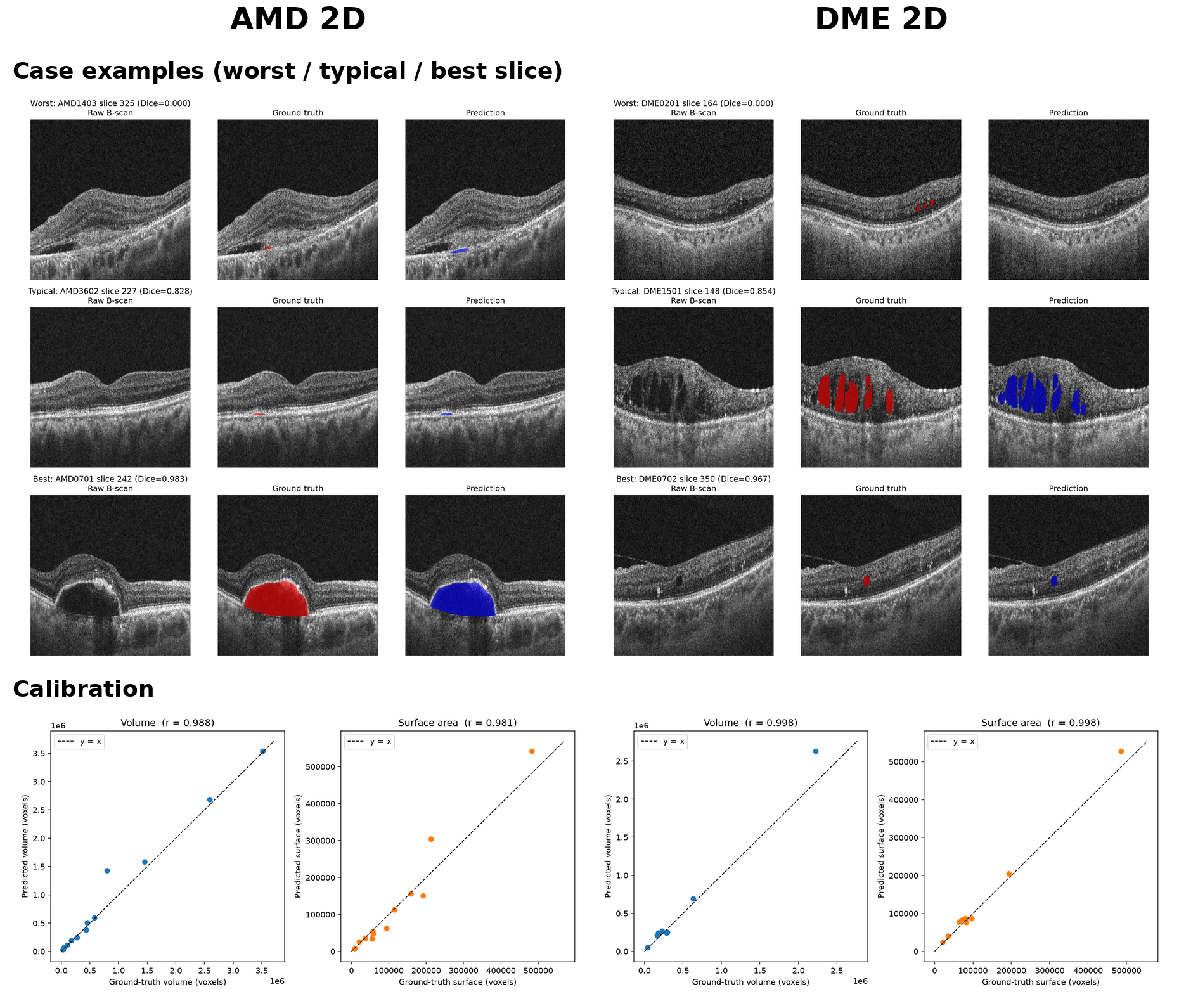}
\caption{AMD 2D and DME 2D case examples (worst-, typical-, and best-case validation slices by Dice) and calibration (predicted vs. ground-truth volume and surface area) for each pipeline's current production configuration.}
\label{fig:combined_2d}
\end{figure}
\FloatBarrier

\subsection{Ablation Studies}\label{subsec:ablation-studies}

Unless otherwise noted, comparisons in this subsection report raw, single-checkpoint Dice at the default 0.5 decision threshold, without test-time augmentation or ensembling --- the isolated effect of the mechanism under test, holding everything else fixed. Test-time augmentation, threshold tuning, and ensemble composition are each substantial contributors in their own right and are treated as separate axes below; the final adopted configurations reported in Table 9 (Final Model Performance) reflect all of these combined.

\subsubsection{Loss Functions}\label{subsubsec:lf}

Tversky-BCE ($\alpha$ = 0.2, $\beta$ = 0.8) was the shared starting loss in all four pipelines; the very first baseline model for AMD 2D used plain Dice loss before switching to Tversky-BCE, which then served as its baseline for every subsequent ablation. Three re-weightings of this baseline were tested: a milder weighting ($\alpha$ = 0.35, $\beta$ = 0.65), a precision-favoring weighting ($\alpha$ = 0.6, $\beta$ = 0.4), and Dice-BCE. Dice-BCE improved standalone 3D checkpoint performance for both datasets (AMD 3D: mean Dice 0.4643 to 0.4793; DME 3D: 0.3894 to 0.4959, its best single 3D checkpoint at that stage of the search) but regressed both 2D pipelines (AMD 2D: no checkpoint cleared the save threshold across 100 epochs; DME 2D: 0.7558 to 0.7379). The milder weighting produced an even sharper divergence within the 3D pipelines alone: a large win for AMD 3D (0.4643 to 0.5494) but a clear regression for DME 3D (0.3894 to 0.2105). The precision-favoring weighting was rejected for AMD in both dimensions, whether tested standalone or as an ensemble candidate, but proved useful for DME 2D specifically. Loss-function sensitivity was therefore dataset- and dimensionality-dependent throughout the project; no single choice dominated across all four pipelines.

However, not every loss variant showed this dataset-dependent pattern. A boundary-aware loss term and a focal Tversky variant were both tested against all four pipelines and rejected uniformly, in every case failing on full-volume Dice or calibration despite occasionally winning a narrower slice-level screening metric.

A per-voxel loss reweighting scheme showed the opposite pattern from small-lesion oversampling (a sampling-side mechanism with the same motivation, covered separately below): it was adopted for DME 3D alone and rejected for the other three pipelines, including two separately-tuned voxel-threshold variants for AMD 3D. Two structurally different mechanisms aimed at the same small-lesion problem therefore generalized in opposite directions --- oversampling helped everywhere it was tried, while loss reweighting helped in exactly one of four cases.

The final pipelines contain four of these loss variants, each producing a measurable gain over the composition it joined: DME 3D's Dice-BCE and per-voxel-reweighted checkpoints (0.7465 to 0.7701, later 0.7963 to 0.7981), AMD 3D's milder-Tversky-plus-Dice-BCE three-member ensemble (0.5967, versus 0.5494 for the strongest individual checkpoint alone), and DME 2D's precision-favoring swap (0.7557 to 0.7658).

\subsubsection{Sampling Strategies}\label{subsubsec:ss}

All 3D checkpoints use balanced positive/negative patch sampling by default. Lesion-guaranteed-only sampling (\verb|pos_ratio=1.0|) was catastrophic for both datasets --- the AMD 3D mean Dice collapsed to 0.12 and DME 3D to 0.10 --- because the resulting all-positive supervision, compounded with the Tversky-BCE recall-heavy weighting ($\alpha$ = 0.2, $\beta$ = 0.8), led to severe over-prediction. An intermediate setting (\verb|pos_ratio = 0.7|), tested only for AMD 3D, also yielded a negative result (mean Dice 0.6021 to 0.5943), confirming that any deviation above the balanced default degrades performance regardless of degree.

Elongation-aware oversampling was a root-cause fix for a large-lesion recall problem traced not to the size of the lesion but to z-extent. This was adopted for both datasets: the AMD 3D ensemble rose from a mean Dice of 0.6477 to 0.6901, and the calibration for DME 3D improved along with Dice rather than trading off.

Small-lesion volume oversampling was tested and adopted in all four pipelines, the most widely useful sampling mechanism found. It raised the AMD 3D mean Dice from 0.710 to 0.738, AMD 2D's full-volume Dice from 0.8214 to 0.8253, and the DME 2D balanced-slice Dice from 0.7688 to 0.7730, with calibration holding or improving in each case; it is also a component of the DME 3D final ensemble. Copy-paste augmentation (lesion instances) was tested for all four pipelines and rejected in every case, consistently losing on full-volume Dice or calibration.

\subsubsection{Patch Depth/Receptive Field}\label{subsubsec:pd}

Increasing patch depth beyond the 64-slice baseline first required solving a training-stability problem: three initial attempts to reach 128-slice depth (at 128xbatch-4, 96xbatch-4, and 128xbatch-2, all at the standard learning rate) collapsed via a BCE-driven cold-start into predicting background everywhere, confirmed by Dice declining in lockstep with the loss. Reducing the initial learning rate 3.3x (1e-4 to 3e-5) resolved this: training remained flat for roughly 17-27 epochs before escaping the plateau and converging normally, a recipe subsequently reused for every other training instability encountered later in the project (including the precision-favoring loss collapse in Loss Functions).

With that fix in place, 128-slice depth was tested with two different losses per dataset. For AMD 3D, a milder-Tversky checkpoint reached a standalone mean Dice of 0.6170 --- the best single AMD 3D checkpoint at that point in the search --- and was added as a fifth ensemble member, resulting in a mean Dice of 0.6193 (t=0.55) with calibration preserved; a second depth-128 checkpoint using the original 0.2/0.8 loss was weaker standalone (0.5087) but was also independently adopted as a further ensemble member, reaching 0.6332. For DME 3D, the milder-Tversky depth-128 checkpoint was weaker standalone (0.7049) than existing members but still improved the ensemble on addition (0.7701 to 0.7829); a second depth-128 checkpoint using Dice-BCE was weaker again standalone (0.5915) but produced a further modest ensemble gain (0.7829 to 0.7860). Therefore, a deeper patch context helped in both datasets at the ensemble level, even when the individual checkpoint was mediocre or weak in isolation.

Two further depth extensions were tested for AMD 3D. Extending to 192 slices reached a standalone mean Dice of 0.5973 and a leave-one-out ensemble gain to 0.6704, but a per-volume check found the large elongated lesion this mechanism was meant to fix and the ensemble's small lesions pulled in opposite directions along the same threshold axis. No single threshold favored both; the mechanism was therefore rejected at every threshold tested. Training directly on the full native volume depth (512 slices, eliminating z-axis sliding-window patching) yielded better results: swapped in for the milder-Tversky checkpoint, it raised the mean Dice from 0.680 to 0.710, with calibration improving on the volume axis (0.963 to 0.968) and unchanged on the surface axis.

At inference time, an increase in the sliding-window overlap fraction from 0.25 to 0.5 was tested for AMD 3D and rejected: the mean Dice moved from 0.5967 to 0.5904, confirming that patch seams were not the actual cause of the fragmentation problem despite roughly doubling the inference compute.

\subsubsection{Architecture Variants}\label{subsubsec:av}

Decoder-level attention gating was rejected for AMD 2D (noise-level, 0.7643 vs. 0.7641), AMD 3D (loses both Dice and pooled Dice against the adopted ensemble), and DME 2D (wins calibration but loses full-volume Dice), but adopted for DME 3D, where a tuned-weight five-way composition raised the mean Dice from 0.8167 to 0.8242 and pooled Dice from 0.8349 to 0.8454 with calibration essentially unchanged.

UNet++ was adopted for both AMD pipelines: for AMD 2D it drove a small-lesion-concentrated gain (Dice 0.4695 to 0.5047 on lesions smaller than 500px, precision and recall improving together) with aggregate Dice roughly flat; for AMD 3D it was individually the weakest standalone checkpoint, yet still a net-positive ensemble member (mean Dice 0.6399 to 0.6422). It was tested independently for DME 2D (Dice 0.7544) but was not adopted.

Both encoder swaps were rejected; EfficientNet-B0 was briefly adopted for AMD 2D before a stale-baseline bug was caught and the decision reversed --- the evaluation had been compared to an already-superseded ensemble composition, and the corrected rerun failed the calibration gate (r\_vol -0.0020); it was also rejected for DME 2D. ResNet-50 was rejected for both AMD 2D and DME 2D, worse on nearly every axis in both cases.

Input-channel augmentation was rejected in every variant tested: DME 3D position channel, DME 3D shadow-darkness channel, and their DME 3D combination all failed to outperform the baseline Dice outright. The AMD 3D position-channel test was the one case where Dice improved, but it regressed on surface calibration; AMD 2D and DME 2D position channel tests were both rejected outright. 

Deep supervision, tested only for AMD 3D, won the balanced-slice sweep but regressed on both pooled Dice and calibration compared to the adopted ensemble and was rejected.

The two-stage cascade/component-verifier network, when evaluated as a general low-confidence component denoiser against the real in-domain val set, cleared the adoption bar (mean Dice 0.8242 to 0.8260, pooled Dice 0.8454 to 0.8467, calibration untouched) and was integrated into the DME 3D production pipeline.

\subsubsection{Augmentation}\label{subsubsec:aug}

Two reduced-recipe variants and a domain-targeted intensity-widening variant were tested against the standard augmentation recipe across the four pipelines.

Removing coarse dropout alone from the AMD 2D recipe was rejected: the resulting checkpoint never beat the full-recipe baseline over 25 epochs, with small-lesion precision, recall, and Dice all regressed.

A more aggressive reduction --- dropping brightness/contrast, Gaussian noise, elastic transform, and coarse dropout entirely, leaving only horizontal flip and rotation to match the reference dataset paper's own minimal recipe --- was tested for all four pipelines and produced a different outcome in each. AMD 2D failed outright, never clearing its save threshold across 25 epochs, confirming the fuller augmentation set is necessary for the model. DME 2D and AMD 3D each won on a Dice-based metric but failed the project's secondary check: a leave-one-out win for DME 2D on the balanced-slice screening metric (0.7730 to 0.7757) reversed on full-volume Dice (0.8243 to 0.8183 at best); a leave-one-out win for AMD 3D on mean Dice (0.7379 to 0.7589) came at a calibration cost (r\_vol 0.9676 to 0.9594). Neither was adopted. DME 3D was the sole clean win: swapped in for the elongation-weighted member, it raised the mean Dice from 0.8024 to 0.8124 with the pooled Dice essentially flat (0.8285 to 0.8277) and improved calibration on both axes (r\_vol 0.9978 to 0.9981, r\_surf 0.9949 to 0.9955), and is retained as a member of the final DME 3D ensemble.

A domain-shift augmentation recipe --- widened brightness/contrast perturbations and a resolution-degrading transform, sized to a measured gap against OLIVES (foreground brightness +48\%, contrast +21\%, native-resolution sharpness roughly 1.8x blurrier) --- was tested across all four pipelines as a candidate for improving out-of-domain generalization and rejected in every case on the same in-domain Dice/calibration bar used throughout this project. The AMD 2D checkpoint narrowly won the balanced-slice screening metric (0.7641 to 0.7671) but fell short of the baseline on full-volume Dice with a mixed calibration result; the DME 2D version lost the screening metric outright (0.7730 to 0.7691) and also every full-volume axis. The full-strength 3D variant triggered gradient-explosion training collapses in both AMD 3D and DME 3D, traced to a missing gradient-clipping safeguard and fixed before retraining. Even after the fix, domain-shift augmentation was rejected for both 3D pipelines: the retrained DME 3D checkpoint reached the highest standalone patch Dice of any DME 3D checkpoint in the project, yet still regressed the ensemble on every axis; a half-strength retry for AMD 3D improved the mean and pooled Dice but regressed on both calibration axes (r\_vol 0.9723 to 0.9543, r\_surf 0.9944 to 0.9930).

Synthetic speckle-noise injection was tested for DME 3D as a candidate fix for wrong-retinal-layer predictions but did not cause an improvement, and remains disabled in production.

\subsubsection{Regularization}\label{subsubsec:reg}

Weight decay was adopted for AMD 2D (full-volume Dice 0.8265 to 0.8282, clean win on every axis at every threshold tested) and for DME 3D, where it remains one of the ensemble's most load-bearing members. For AMD 3D, it was adopted once, as a fourth member of an early ensemble composition (mean Dice 0.5967 to 0.6018), but that composition was fully superseded by later ensemble refreshes; a later retry matching the exact loss pairing for DME was tested against the current 5-way baseline and rejected (wins raw Dice, loses pooled Dice and both calibration axes). It was also rejected for DME 2D (wins the balanced-slice sweep but loses full-volume Dice at every threshold).

LR-decay removal did not improve standalone performance for AMD 2D or DME 2D --- mild negative results in both cases, never clearing the respective baselines of either model. The AMD 2D variant was not carried forward into that pipeline's ensemble; the DME 2D variant, despite the same standalone loss, was still retained as a diversity-contributing member of that pipeline's final ensemble (see Ensemble Composition).

Extending training from 100 to 120 epochs was tested for both AMD 3D and DME 3D and found no benefit in either case --- both checkpoints had genuinely converged, with the AMD 3D extension actually performing slightly worse on full re-evaluation despite a nominal training-metric improvement.

\subsubsection{Post-Processing}\label{subsubsec:pp}

TTA was extended with a third pass (vertical flip in 2D, z-axis flip in 3D) beyond the shared horizontal-flip baseline. The z-flip variant was adopted for DME 3D (mean Dice 0.7981 to 0.8024, pooled Dice 0.8225 to 0.8285, calibration essentially unchanged), but rejected for AMD 3D due to calibration cost. The vertical-flip variant was rejected for both 2D pipelines --- the mean Dice of AMD 2D dropped from 0.8253 to 0.8215 despite calibration improving, and DME 2D experienced no real gain; the change was therefore not worth the permanent inference cost.

Decision-threshold tuning is adopted universally; every final recipe uses a per-pipeline tuned threshold rather than the naive 0.5 default. The clearest case is DME 3D's early single-checkpoint recipe, where sweeping the threshold on TTA-averaged probabilities took the mean Dice from 0.4999 to 0.7205 (t=0.95) and directly corrected a systematic volume over-prediction that morphological erosion alone could not reach.

The AMD 3D early post-processing chain --- 3D binary erosion (tested on the single best checkpoint before ensembling began), closing/hole-filling, and minimum-component-size filtering --- was adopted in sequence against early ensemble compositions (closing/hole-fill: mean Dice 0.5967 to 0.5980; min-size at 20,000 voxels: 0.5980 to 0.6092), with largest-connected-component filtering rejected outright because it discarded too much signal on small-lesion volumes. None of this carried forward once the ensemble was rebuilt around the depth128/UNet++-era checkpoints --- the final AMD 3D recipe runs with no post-processing. Closing/hole-filling was also tested for AMD 2D (negligible gain, not adopted) and DME 3D (all three metrics moved slightly down, not adopted); multi-focal lesion structure of DME leaves little for the closing operation to address.

The edge-aware guided-filter boundary refinement was tested only for AMD 3D and rejected; it improved the mean Dice but regressed on the pooled Dice and both calibration axes.

Three geometry/image-based filters targeting DME 3D out-of-domain false positives (discussed further in OLIVES Validation, below) were rejected: the fill-ratio and axial-elongation filters collapsed Dice catastrophically at any threshold that removed components, while the shadow-overlap filter never exceeded the baseline Dice at any threshold, with calibration essentially untouched either way.

The plausibility filter evolved through three versions, all for DME 3D. The original intensity-and-texture rule cost a small, accepted amount of in-domain Dice (0.8245 to 0.8218) in exchange for suppressing most of an OLIVES artifact pattern; a band-position extension followed. The current version adds a voxel-count size ceiling exempting large components from the suppression rule entirely, motivated by evidence that real in-domain lesions rarely trip the rule regardless of size. This was a genuine in-domain gain as well, not just an out-of-domain fix, outperforming both the unfiltered baseline and the original filter's small cost. It is the current production configuration.

\subsubsection{Ensemble Composition}\label{subsubsec:ec}

Combining independently trained checkpoints via probability averaging has been the single largest lever in the project, consistently dwarfing the gains from any individual training-side mechanism. It was originally motivated by demonstrated run-to-run training noise (identical AMD 3D configurations ranged from mean Dice 0.42 to 0.62 across repeated runs); averaging multiple checkpoints' predictions reduced that variance directly. The magnitude has stayed large as the project matured: the best reproducible single checkpoint for AMD 3D reached a mean Dice of 0.5494, against a current 5-way ensemble at 0.738; the best single checkpoint for DME 3D reached a mean Dice of 0.7417, against a current ensemble at 0.8247.

Leave-one-out composition searching became the standard methodology for nearly every adoption decision in this project and repeatedly overturned the intuitive assumption that only individually strong checkpoints are worth including. The AMD 3D UNet++ member was the weakest standalone checkpoint found in an entire session, yet net-positive once ensembled; the milder-Tversky checkpoint for DME 3D was individually the worst of three checkpoints, yet still improved the ensemble by contributing diversity. 

Weighted ensembling via coordinate-ascent search was adopted for both final 3D pipelines (the current weights for DME 3D range from 0.098 to 0.39 across its five members; AMD 3D ranges from 0.089 to 0.444 across its five members) and the final ensembles for the 2D pipelines use uniform averaging. A DME 2D coordinate-ascent search (tested as a candidate, not adopted) once found an apparent gain by zeroing out two of four original members' weights entirely, which appeared to be a real improvement on the balanced-slice screening metric but failed the full-volume calibration check outright --- a sign the search had overfit the screening metric rather than found a real improvement.

\subsection{OLIVES validation}\label{subsec2}

The OLIVES dataset is an independent cohort, collected on different scanners under different clinical protocols than the Huang et al. dataset. It serves as a reference for the generalizability of these models beyond their in-domain training data. Because OLIVES has no lesion-level ground truth, this work is based on the proxy metrics described in Methods (biomarker AUROC, CST/volume correlation, cross-visit concordance) rather than Dice, and on qualitative visual review where the clinical labels of OLIVES do not reach lesion-level detail.

\begin{figure}[h!]
\centering
\includegraphics[width=0.6\textwidth]{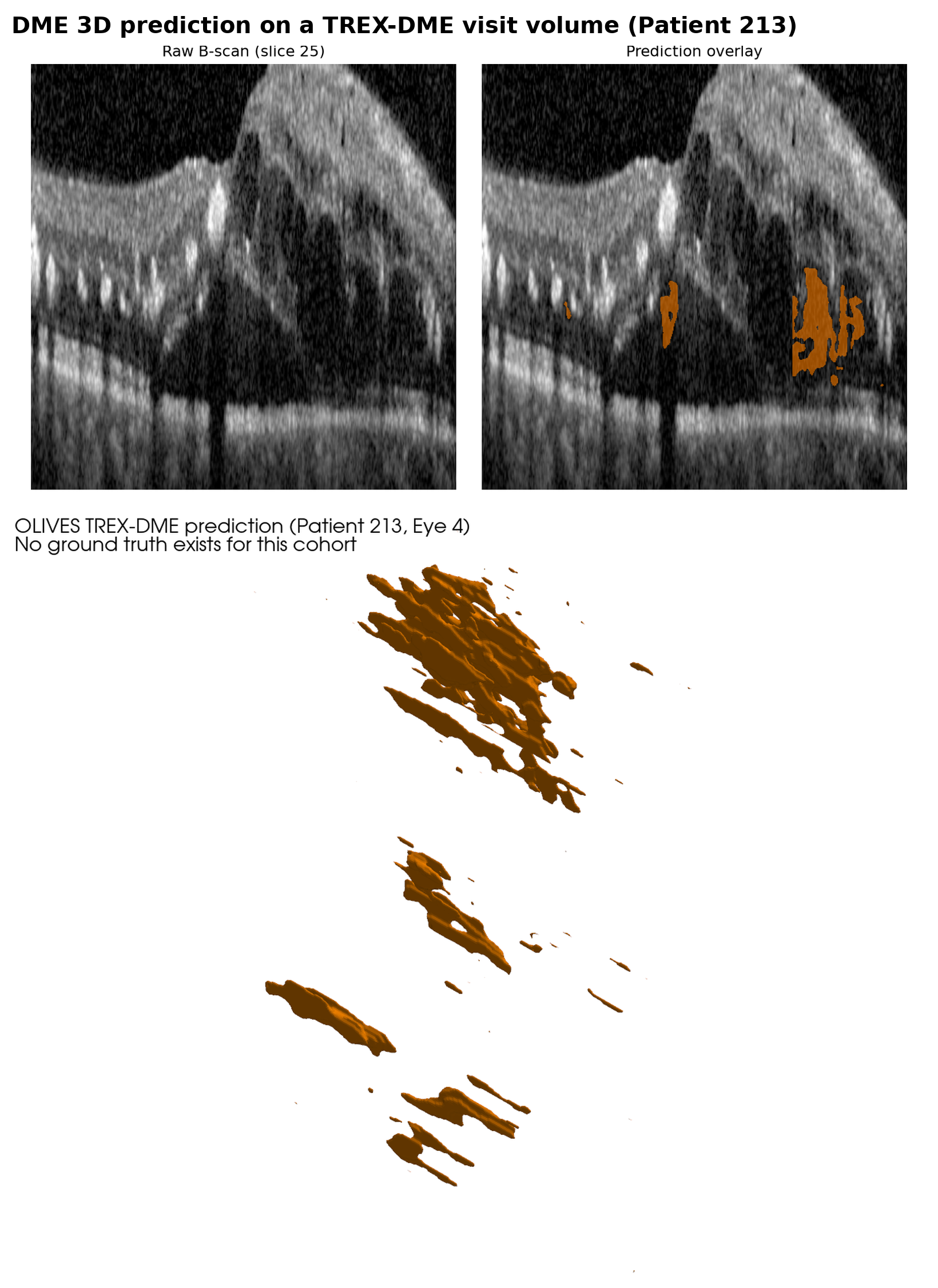}
\caption{DME 3D's full predicted volume for one TREX-DME visit (Patient 213), current production configuration, rendered as a 3D surface --- shown alongside a representative raw B-scan from the same visit for anatomical context. No ground truth exists for this cohort. The confirmed shadow-artifact patients (203, 204) were deliberately excluded from selection, since they are discussed separately below.}
\label{fig:olives_3d}
\end{figure}

\begin{figure}[h!]
\centering
\includegraphics[width=0.85\textwidth]{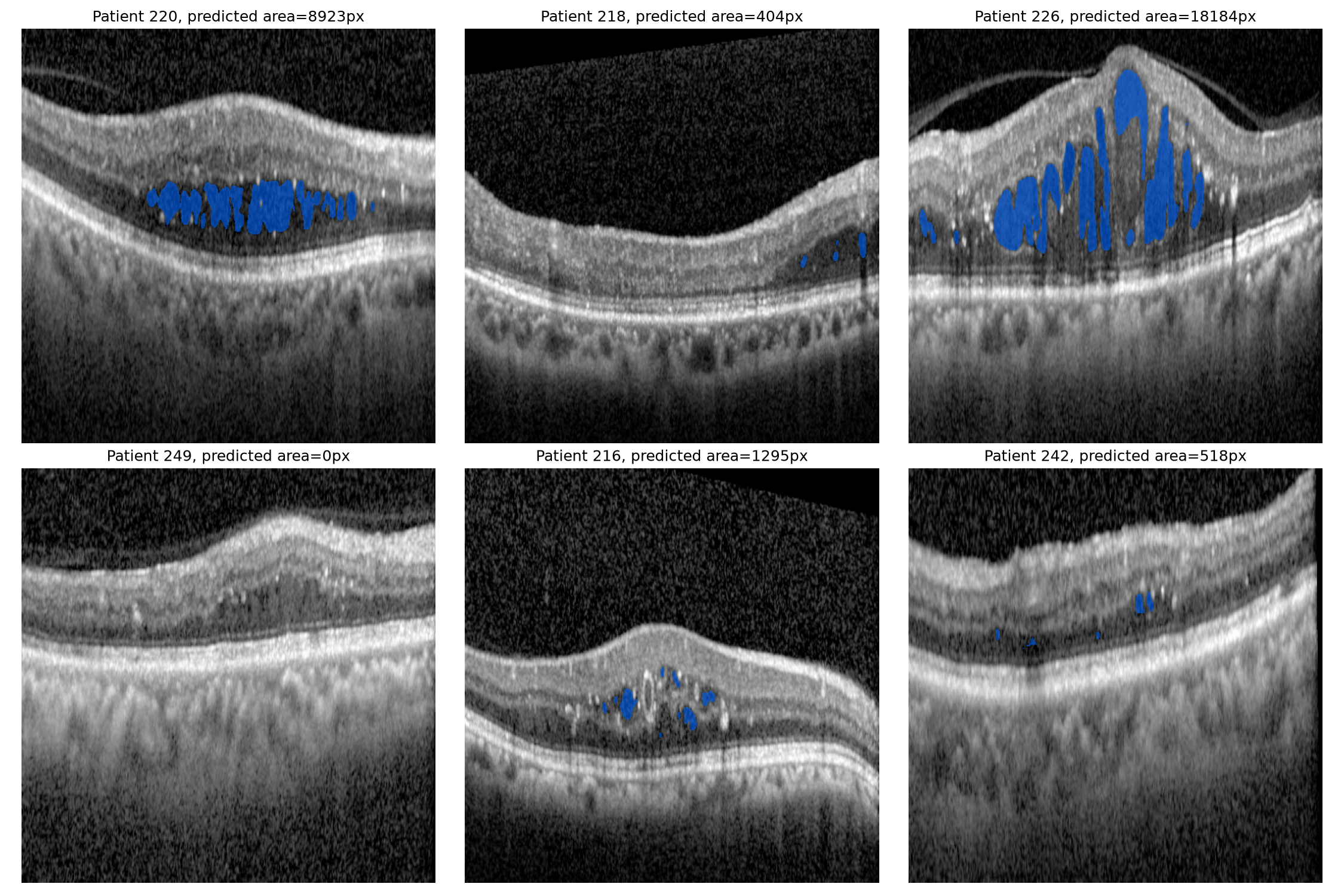}
\caption{DME 2D predictions on six TREX-DME IRF-positive B-scans, randomly sampled across six different patients, current production ensemble and envelope guardrail. Since OLIVES carries no lesion-level ground truth, these overlays are the primary visual evidence for this section.}
\label{fig:olives_2d}
\end{figure}

Evaluated on TREX-DME (the disease-matched cohort), the DME 3D pipeline reaches an IRF AUROC of 0.7282, r(CST, pred\_vol) = 0.5706 (p<0.0001, n = 76), and cross-visit concordance of 83.3\% (30/36); the DME 2D pipeline reaches an IRF AUROC of 0.9072, r(CST, pred\_vol) = 0.1883 (p = 0.45, n.s., n = 18), and cross-visit concordance of 82.4\% (14/17). The biomarker AUROC for DME 2D is notably stronger than for DME 3D despite the weaker patch-level architecture, though its CST correlation is weak and not statistically significant on a much smaller matched-visit sample; the two pipelines otherwise generalize comparably well on concordance. Since there is no ground truth accompanying the numbers above, Figures~\ref{fig:olives_3d} and \ref{fig:olives_2d} show representative predictions for both pipelines directly.

As a check on the TREX-DME restriction described in Methods, the same AUROC metrics were also computed on PRIME (whose patients were enrolled without DME). Both pipelines performed markedly worse there: DME 3D reaches an IRF AUROC of 0.5583 (vs. 0.7282 on TREX-DME), and DME 2D reaches an IRF AUROC of 0.8285 (vs. 0.9072), consistent with the disease-mismatch rationale in both cases. 

The envelope guardrail was adopted because it fixed a visible, clinically obvious failure mode that the aggregate AUROC could not properly describe: without it, predictions regularly extended into anatomically implausible regions outside the retinal envelope, a failure invisible to a slice-level AUROC computed only against biomarker presence or absence.

A bug in the guardrail itself was later discovered and fixed. The original ILM-tracing implementation used a dynamic-programming shortest path with a smoothness penalty that resisted following steep local tissue elevation --- on B-scans with a pronounced DME edema dome, the traced line cut straight across the base of the dome rather than following the elevated surface, misclassifying fluid pockets sitting inside the dome as "above the retina" and erasing them. Checked against 301 TREX-DME IRF-positive slices, this guardrail cut more than 50\% of the predicted area in 14.6\% of the slices and erased predictions in 7.6\%, including one large, unambiguous cystoid fluid pocket completely removed by a single sharp jump in the traced line at the edge of the dome. Seven iterations of the line-tracing parameters were tested before arriving at an asymmetric-smoothness fix (a low smoothness weight for the ILM line, which only needs to find one transition; a higher weight for the RPE line, which can otherwise jitter between multiple plausible transitions inside a thickened dome). The adopted fix cut total-erasure cases roughly four-fold (7.6\% to 2.0\%) while leaving average clipping essentially unchanged, resolving the specific catastrophic failure mode without meaningfully changing the behavior of the guardrail.

Two OLIVES patients (203, 204) show a persistent thin-streak false-positive pattern, and resolving it became a ten-mechanism investigation, all but the last unsuccessful. Three geometry/image-based filters (fill-ratio, axial elongation, shadow-column overlap; covered in Post-Processing) collapsed the in-domain Dice at any threshold aggressive enough to touch the artifact. Six further characterization attempts also failed to separate the artifact from real components: a counterfactual shadow-removal intervention left the confirmed streaks unchanged; a cross-visit persistence check found no separation; a multivariate classifier over seven features was reverted after proving confidently wrong on the largest components; an en-face vessel-line continuity check found its signature was common to large components generally; and a graded-suppression score and an ensemble-disagreement signal were both separable only in aggregate, not on pertinent components.

The eventual fix abandoned discrimination entirely: since real large in-domain lesions seldom satisfy the suppression criteria, the current plausibility filter simply exempts any component above a voxel-count ceiling from suppression. This recovers the large components that the blunt rule was also catching, improving both the in-domain Dice and the OLIVES proxy metrics reported above.

The two-stage cascade --- adopted into the general production pipeline for its in-domain denoising benefit (see Architecture Variants) --- was also tested directly against patients 203/204 as a candidate fix for this artifact. It does not solve it: the artifacts survive cascade suppression at essentially unchanged size in both patients, even though the cascade suppresses over half of all flagged components elsewhere in the same volumes. Its production adoption stands on the separate, unrelated general-denoising result, not on any success against this specific artifact.

Taken together, these results characterize out-of-domain work of this project as an open validation effort; the proxy metrics show real signal, but they remain indirect substitutes for ground truth the model was never given, and the artifact investigation above shows how much of that signal can still be muddied by domain-shift failure modes that a single suppression rule can only partially separate from real pathology. Establishing a more direct, mechanistically grounded relationship between what the model predicts and the clinical biomarkers provided by OLIVES remains an unfinished work.

\subsection{PED Analysis}\label{subsec2}

Serous PED, the AMD-relevant biomarker OLIVES provides, is a rare comorbid finding within the DME-focused TREX-DME cohort --- there are only 10 positive slices in the entire dataset (one patient), too few for a formal AUROC and restricted to qualitative review. All 10 positive slices come from a single patient with substantial co-occurring pathology --- every slice is also positive for preretinal tissue/hemorrhage and fully attached vitreous face, and 6/10 are also positive for DME-related fluid (DRT/ME, IRF). Since AMD 2D was trained only to detect PED, recovered near-RPE activation cannot be cleanly attributed to serous PED alone; it may also reflect these co-located findings, which the model was never trained to distinguish from PED. OLIVES grades only the serous subtype specifically, while this project's training masks may include other PED subtypes; the label is therefore a strong but imperfect match rather than a guaranteed one-to-one correspondence. AMD 3D fails to detect the label at the standard operating threshold (mean 0.5px predicted area across the 10 known-positive scans, 9/10 exactly 0px); a relative/local-threshold follow-up produced suspiciously uniform \~166-167px areas regardless of scan content, diagnosed as a method artifact rather than real detection and not pursued further.

\begin{figure}[h!]
\centering
\includegraphics[width=0.7\textwidth]{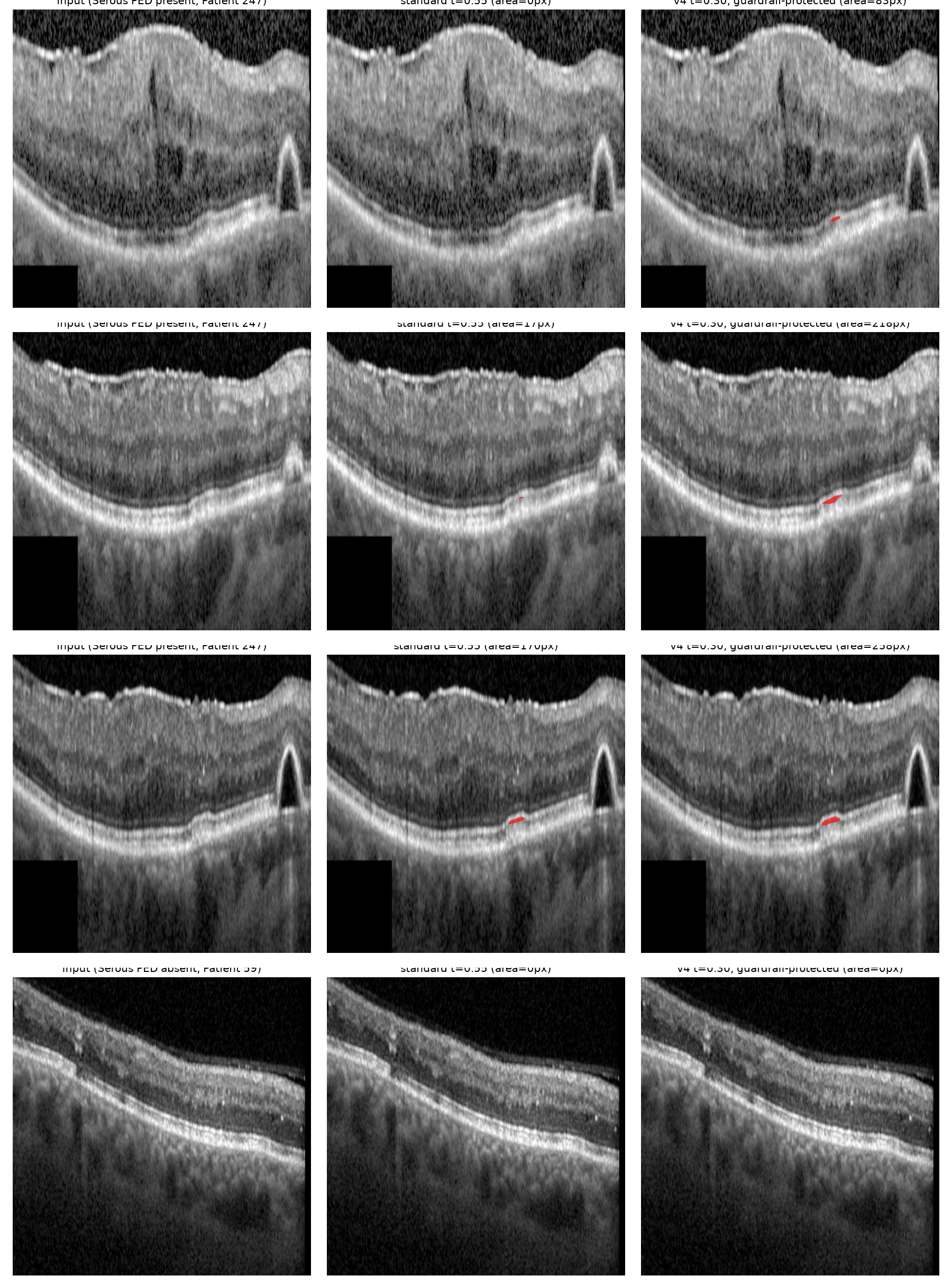}
\caption{AMD 2D raw probability response on serous PED slices, comparing the standard operating threshold (t=0.55) against a lowered, guardrail-protected threshold (t=0.30). Top three rows: PED-positive slices from Patient 247, showing recovered signal at the correct near-RPE location, including one case entirely missed at the standard threshold (0px) but recovered at the lowered one. Bottom row: a PED-negative slice at the same lowered threshold, for contrast --- confirming the recovered signal is anatomically specific, not generic threshold noise.}
\label{fig:ped_overlays}
\end{figure}

AMD 2D performed better, but was still weak and inconsistent at the standard threshold (6/10 positive slices nonzero, ranging 16-169px; 4/10 exactly zero). Several follow-up mechanisms --- a guardrail line-detector variant, a training-representation shape analysis, and a targeted brightness/contrast normalization matching the measured photometric domain gap of OLIVES --- were tested as fixes, and each yielded a negative result. A closer look at the raw, unthresholded probability maps then found missed signal: several slices show a weak but genuine elevated probability (0.2--0.4) at the correct near-RPE dome location, sitting below the standard threshold. Sweeping the threshold downward within the already-adopted RPE-proximity guardrail region recovered coherent signal in 8/10 positive slices, confirmed against an explicit negative control (15 PED-negative slices stayed near-zero, averaging 1.7px, under the same lowered threshold) --- genuine anatomical specificity, not generic noise inflation from a lower cutoff. Figure~\ref{fig:ped_overlays} shows this directly, including a case missed at the standard threshold and the negative-control comparison. This is a qualitative single-patient finding (n=10 positive slices, no formal AUROC possible) and is not proposed as a change to the AMD 2D production operating threshold used elsewhere. 

\section{Discussion}\label{sec12}

All four pipelines reached a similar operating point --- Dice scores in the 0.76--0.82 range with strong volumetric and surface calibration --- although AMD 3D trails the other three by a modest margin (mean Dice 0.7623, pooled 0.7573, r\_vol 0.9723, r\_surf 0.9927, vs. pooled Dice 0.844--0.849 and r\_vol $\geq$ 0.988 for AMD 2D, DME 2D, and DME 3D). AMD 3D underwent the same ablation process as every other pipeline, but its smaller, more variable 3D lesion morphology ultimately proved to be the most difficult segmentation task.

The ablation results illustrate that loss-function and architecture changes produced smaller, more pipeline-specific gains, while ensemble composition was consistently the largest lever. A second pattern recurred often enough to be worth stating as a general point about evaluating small-lesion segmentation: several mechanisms that won on a balanced or positive-slice evaluation sweep lost once checked against full-volume Dice and calibration. Screening changes only on lesion-present slices risks adopting a change that appears to be an improvement, but degrades full-volume performance.

The OLIVES results reveal a more modest picture than the in-domain numbers above. In the disease-matched TREX-DME cohort, DME 3D reaches a biomarker AUROC of 0.73 (IRF), a Pearson correlation of 0.57 (p<0.0001) between the predicted volume of the lesion and CST, and a directional concordance of 83.3\% between the change in consecutive-visit volume and the change in CST; DME 2D reaches a higher AUROC (0.91 IRF) but a much weaker, non-significant CST correlation on a smaller eligible sample. Performance with the PRIME cohort, where DME is not part of enrollment criteria, is consistently weaker across both pipelines. This pattern is clinically reassuring, as it indicates that the models are detecting a disease-specific signal rather than a generic imaging artifact. Together, these results make the case for external proxy-metric validation: strong in-domain Dice does not by itself establish that a model will behave sensibly on unseen scanners, protocols, or patient populations.

The PED analysis illustrates the limits of this proxy-only approach. AMD 2D shows anatomically-specific signal for serous PED that the standard operating threshold does not surface, while AMD 3D shows essentially none --- but the entire finding rests on 10 positive slices from a single patient, qualitative by necessity since OLIVES carries no lesion-level ground truth. Where the biomarker AUROC and concordance checks above draw on tens to hundreds of slices and dozens of visits, PED is a reminder that this same methodology is constrained by the extent of the available data.

The broader clinical motivation for this work is the longitudinal monitoring of lesion burden alongside biomarkers to inform treatment decisions in AMD and DME. The four pipelines built and systematically ablated here show strong in-domain performance in favor of this goal, and the OLIVES testing extends this result cautiously outside the training distribution: predicted volume tracks CST across visits often enough to suggest that the same signal survives, at least for the disease-matched cohort. Together, these results are evidence for automated lesion-burden tracking as a clinical tool, not a validation of it --- the proxy metrics used throughout this work stand in for clinical outcomes because no lesion-level ground truth or outcome data were available to test prediction accuracy directly.

This work has several limitations that motivate future work. No held-out test set was reserved from the training data; every mechanism adoption decision and the final reported numbers were evaluated against the same 13 AMD / 9 DME volumes. Therefore, these results should be read as development-set performance under a fixed decision rule rather than as an independent generalization estimate. Validating these models against volumes untouched by any tuning decision remains an open step. All training data come from a single source (Huang et al.) collected on one scanner under one protocol, and OLIVES itself has no lesion-level ground truth, so every external result here is a proxy rather than a direct accuracy measurement; a larger, more diverse external cohort with at least some lesion-level annotation would allow future work to move these proxy metrics toward a direct accuracy check. The underperformance of AMD 3D relative to the other three pipelines is documented here but not resolved, and remains an open target for further investigation. The PED finding requires an adequately powered dataset before it can be considered more than a qualitative observation. Longer-term, the concordance and calibration metrics developed for OLIVES underscore the need for longitudinal, per-patient monitoring of lesion burden aimed at strengthening the relationship between the model predictions and the clinical biomarkers they are intended to track. 

\section{Conclusion}\label{sec13}

This work aimed to determine whether deep-learning-based lesion segmentation can reliably characterize AMD and DME lesions on OCT images, both within the distribution in which it was trained and in clinical data the model never saw during training. Four pipelines (2D and 3D variants for each disease) were built and systematically ablated, achieving Dice scores of 0.76--0.82 with strong volumetric and surface calibration. In OLIVES, an external clinical cohort with no lesion-level ground truth, predictions tracked real biomarkers and clinical measurements such as CST outside the training distribution, though they were generally attenuated relative to in-domain performance. Beyond the models themselves, this work's contribution lies in its methodology: a full-volume, calibration-aware adoption standard that caught mechanisms an ordinary slice-level sweep would have kept, and a proxy-metric framework that makes external validation possible even where no ground truth exists to check against directly. What remains is to close the gap between that proxy-metric evidence and clinical validation, with the ultimate goal of testing these models directly against clinical outcomes and strengthening the relationship between model predictions and the biomarkers they are meant to track.

\section*{Declarations}

\begin{itemize}
\item Funding: The authors received no funding for this work.
\item Conflict of interest/Competing interests: The authors declare no competing interests.
\item Ethics approval and consent to participate: This study used publicly available, de-identified imaging datasets (Huang et al.; OLIVES). Data from human subjects were not collected as part of this work.
\item Consent for publication: Not applicable.
\item Data availability: The training dataset (Huang et al.) is publicly available at https://doi.org/10.6084/m9.figshare.30582035. The external validation cohort (OLIVES) is publicly available \citep{prabhushankar2022olives}.
\item Materials availability: Not applicable.
\item Code availability: Code is available from the corresponding author upon reasonable request.
\item Author contribution: The first author performed all model development, experimentation, analysis, and manuscript writing, with supervision from Dr. Zhihao Zhao.
\end{itemize}

\bibliography{sn-bibliography}

\end{document}